%% file: 00_main.tex
\documentclass[letterpaper, 10 pt, conference]{ieeeconf}
\IEEEoverridecommandlockouts                               
\usepackage{breqn}
\usepackage{cuted}
\usepackage{xcolor}
\usepackage{capt-of}
\usepackage{amsfonts}
\usepackage{hyperref}
\usepackage{multirow}
\usepackage{mathtools}
\usepackage{dblfloatfix}
\usepackage{makecell}
\usepackage{amsmath}
\usepackage{amsfonts}
\usepackage{amssymb}
\usepackage{graphicx}
\usepackage{algorithm}
\usepackage{algpseudocode}
\usepackage{booktabs}

\usepackage[font=small, labelfont=bf]{caption}

\usepackage{amssymb}

\usepackage{fancyhdr} 
\renewcommand{\arraystretch}{1.2}

\definecolor{pink}{RGB}{255, 192, 203}

\newcommand{\del}[1]{}

\usepackage[T1]{fontenc}      
\usepackage[utf8]{inputenc}   
\usepackage[vietnamese]{babel}

\title{\LARGE \bf $\text{T}^2$-Nav: Algebraic-Topology-Aware Temporal Graph Memory and Loop Detection for Zero-Shot Visual Navigation}

\author{
    Quang-Anh N. D.$^{1,4}$ \thanks{$^1$ International School, Vietnam National University, Hanoi, Vietnam.}%
    \and Duc Pham Minh$^{2,4}$\thanks{$^2$ Hanoi University of Science, Vietnam National University, Hanoi, Vietnam.}%
    \and Minh-Anh Nguyen$^1$\thanks{$^3$ Hanoi University of Science and Technology, Hanoi, Vietnam.}%
    \and Tung Duy Doan$^3$\thanks{$^4$ *Cognitive Robotics Lab, Department of Electrical Engineering and Computer Science, 
    University of Arkansas, Fayetteville, AR, USA.}%
    \and Tuan Dang*$^4$\thanks{(Corresponding author: \texttt{tuand@uark.edu})}%
}

\begin{document}

\maketitle 

\thispagestyle{empty}
\pagestyle{empty}

\begin{abstract}
Deploying autonomous agents in real-world environments is challenging, particularly for navigation, where systems must adapt to situations they haven’t encountered before. Traditional learning approaches require substantial amounts of data, constant tuning, and, sometimes, starting over for each new task. That makes them hard to scale and not very flexible. Recent breakthroughs in foundation models, such as large language models and vision-language models, enable systems to attempt new navigation tasks without requiring additional training. However, many of these methods only work with specific input types, employ relatively basic reasoning, and fail to fully exploit the details they observe or the structure of the spaces. Here, we introduce $\text{T}^2$-Nav, a zero-shot navigation system that integrates heterogeneous data and employs graph-based reasoning. By directly incorporating visual information into the graph and matching it to the environment, our approach enables the system to strike a good balance between exploration and goal attainment. This strategy allows robust obstacle avoidance, reliable loop closure detection, and efficient path planning while eliminating redundant exploration patterns. The system demonstrates flexibility by handling goals specified using reference images of target object instances, making it particularly suitable for scenarios in which agents must navigate to visually similar yet spatially distinct instances. Experiments demonstrate that our approach is efficient and adapts well to unknown environments, moving toward practical zero-shot instance-image navigation capabilities. Our source code is available at \href{https://github.com/cogniboticslab/t2nav}{https://github.com/cogniboticslab/t2nav}.


\end{abstract}

\input{01_introduction}
\input{02_related_work}
\input{03_system}
\input{04_methodology}

\input{05_evaluation}
\input{06_conclusions}

\bibliographystyle{IEEEtran}
\bibliography{IEEEabrv, 09_references}

\end{document}

%% file: 01_introduction.tex
\section{Introduction}
Deploying autonomous agents in real-world environments poses a significant challenge in robotics, with navigation being a core requirement for applications such as service robots and exploration systems \cite{ZhengYSZWZZC25}. While researchers have made progress in lab settings \cite{krantz2022instance, bhattacharya2015persistent, pokorny2016high, kwon2021visual, cai2024dgmem}, real-world deployment remains limited by systems' inability to operate in unseen environments without massive, task-specific training. This demands zero-shot navigation methods that generalize instantly without extensive data or adaptations. Instance-image navigation (IIN) is one of the most challenging problems, in which systems must locate specific object instances using only reference images as guidance. Unlike traditional object-category navigation that relies on semantic class labels, IIN demands fine-grained visual understanding to distinguish between visually similar objects and identify the exact target instance \cite{lei2024gaussnav}. In particular, real-world environments exhibit substantial variability in lighting conditions, object arrangements, architectural layouts, and dynamic elements; the same object instance may appear dramatically different from different viewpoints, under varying lighting conditions, and with partial occlusions \cite{Xiao2025}.

\begin{figure}[t]
    \centering
    \includegraphics[width=1\linewidth]{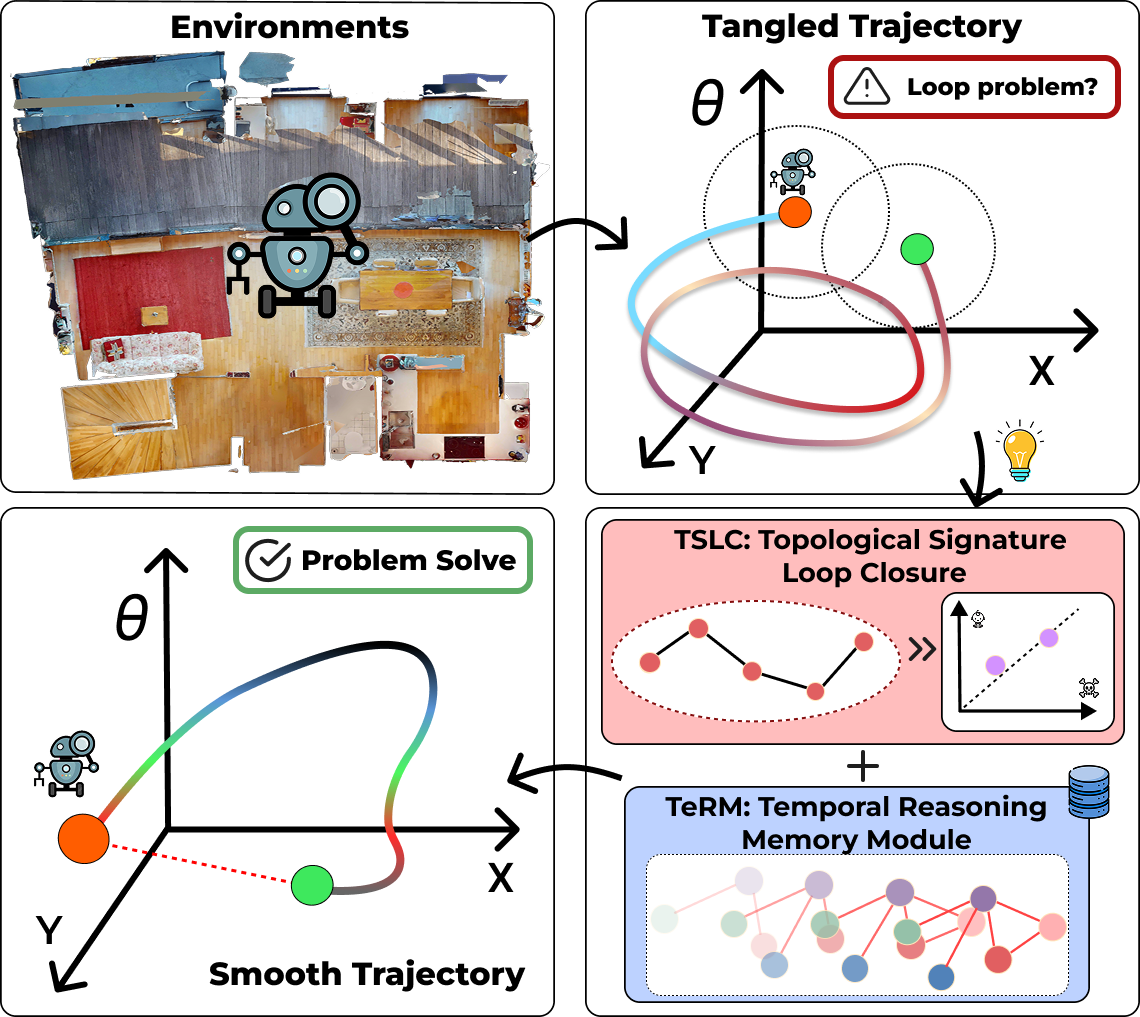}
    \caption{Conceptual overview of T$^2$-Nav. The framework addresses loop-closure problems in which the agent initially becomes trapped in repetitive exploration patterns. To address this issue, we propose two novel modules: (1) TeRM maintains cross-temporal object relationships, and (2) scene dynamics are explored through graph-based reasoning for TSLC to employ persistent homology, enabling the detection and avoidance of navigation loops via topological invariants of agent trajectories, thereby facilitating a robust zero-shot visual navigation system.
    }
    \label{fig:concept}
\end{figure}

The IIN's practical importance spans numerous applications, from service robots tasked with retrieving specific items in households to warehouse automation systems that must locate particular instances of products among thousands of similar items. These applications demand navigation systems capable of understanding visual nuances that distinguish one instance from another, even when they belong to the same semantic category, while also navigating to semantically meaningful locations (\textit{e.g.}, "kitchen table", "office desk") rather than simple coordinate-based targets \cite{SUN2025103135}. Current approaches to address these challenges fall into two main categories: traditional supervised learning methods and emerging foundation model-based approaches. While recent advances in deep learning have shown promising results in controlled scenarios, the gap between laboratory performance and real-world deployment remains substantial.

Traditional navigation methods based on supervised learning face several critical limitations in visual navigation tasks. First, these methods require millions of training samples, which in turn require substantial computational resources \cite{Wijmans2020DDPPO}. Additionally, they perform poorly in environments that differ from those for which they were trained \cite{Anderson2018VLN}. Furthermore, they cannot handle new objects or goals that were not included in their training data \cite{Ramakrishnan_2022_CVPR}. Most importantly, they must be retrained from scratch for each new task. On the other hand, foundation models such as Large Language Models (LLMs) and Vision-Language Models (VLMs) exhibit zero-shot potential, with early successes in navigation integration. Early research has shown promising results for zero-shot in navigation tasks \textit{e.g.} CLIP-Nav \cite{Dorbala2022CLIPNav} for vision-language tasks, and VLFM \cite{pmlr-v205-shah23b} has been tested on real robots. However, methods such as ZSON \cite{Majumdar2020VLNWebText} and SG-Nav \cite{Yin2024SGNav} are modality-specific, lack universality, and employ simplistic reasoning, leading to inefficient exploration in complex spaces.

The key gap is unified frameworks for diverse visual goals, while also combining robust spatial and semantic information, recent developments in scene representation and reasoning in foundation models, including graph-based \cite{yin2025unigoal}, and multi-modal approaches \cite{pmlr-v162-li22n} enable universal methods for zero-shot navigation. However, existing methods have not fully utilized the visual information in these representations or graph-structured approaches for robust navigation planning. Particularly, existing training-free approaches suffer from two critical limitations: (1) inability to detect complex loop patterns beyond simple geometric proximity, leading to redundant exploration, and (2) lack of temporal coherence in scene representation, causing inconsistent goal recognition across viewpoints. We address these gaps using a combination of temporal graph memory and topological loop detection, both of which operate without any learned parameters. While recent graph-based approaches, such as UniGoal~\cite{yin2025unigoal}, provide universal goal handling through multi-stage matching, they remain vulnerable to navigation loops and lack mechanisms for temporal reasoning about scene evolution.

In this paper, we propose $\text{T}^2$-Nav, a novel framework to address the problems above, as described in Figure~\ref{fig:concept}. Which enhances graph representation and analysis by leveraging visual properties and graph matching to balance exploration and navigation, while mitigating issues such as obstacles, loop closure, and redundant paths. In summary, we make the following contributions:

\begin{itemize}
    \item \textbf{Temporal Graph Memory Networks (TeRM):} We introduce a novel temporal reasoning framework that maintains cross-temporal edges between scene graphs, capturing object permanence and dynamics to address inconsistent goal recognition across viewpoints.
    \item \textbf{Topological Signatures for Loop Closure (TSLC):} We propose an application of persistent homology to training-free navigation, utilizing topological invariants to detect complex loop patterns beyond simple geometric proximity, significantly reducing redundant exploration.
\end{itemize}

\begin{figure*}[ht!]
    \centering
    \includegraphics[width=\linewidth]{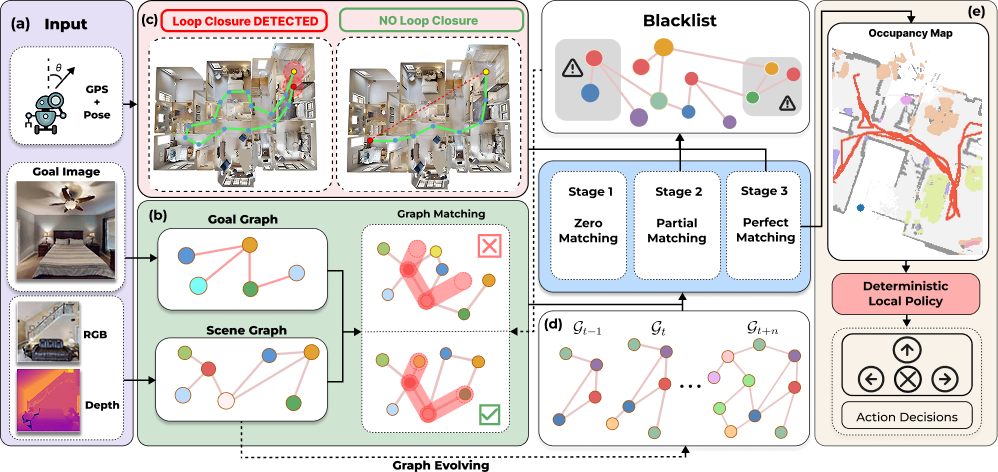}
        \caption{Overview of $\text{T}^2$-Nav system. (a) Multi-modal inputs with GPS, pose, and goal image specifying the target instance, and RGBD input to construct a dynamic scene graph. (b) Graph Processing with RGBD images performs scene-goal matching to identify potential target instances, with \textcolor{green}{\checkmark}/\textcolor{red}{$\times$} indicating success/failure. (c) Loop closure detection maintains a blacklist of locations. (d) TeRM provides temporal consistency for robust instance tracking. (d) The TeRM module maintains temporal consistency across consecutive scene graphs for robust instance tracking across varying viewpoints and environmental conditions. (e) The navigation pipeline generates occupancy maps, applies a deterministic local policy for obstacle avoidance, and outputs action decisions.
        }
    \label{fig:pipe}
\end{figure*}

%% file: 02_related_work.tex
\section{Related Work}
\subsection{Learning-Based Visual Navigation}
Instance-image navigation (IIN) is a challenging task in embodied AI, where agents locate specific object instances in unseen environments using egocentric visual observations and goal specifications as reference images~\cite{krantz2022instance}. Approaches have evolved from geometry-based methods relying on explicit maps or graphs for planning~\cite{bhattacharya2015persistent, pokorny2016high} to learning-based paradigms using Reinforcement Learning (RL) and Imitation Learning (IL)~\cite{kwon2021visual, cai2024dgmem}. While geometry-based techniques provide precision, they often lack semantic priors for adaptability~\cite{krantz2022instance}. RL/IL methods learn flexible policies from interactions, as in works on target-driven navigation~\cite{Zhu2017TargetDriven}. However, these require extensive interactions and struggle with generalization to unseen environments. Our framework leverages mathematical principles and foundation-model representations to enable zero-shot navigation without parameter learning or adaptation, yielding a training-free approach.

\subsection{Zero-shot Visual Navigation}
Recent advancements in zero-shot visual navigation leverage foundation models like Large Language Models (LLMs) and Vision-Language Models (VLMs) for generalization without environment-specific training~\cite{Feng2025SurveyLLM}, with works like~\cite{yin2025unigoal} demonstrating LLMs' zero-shot planning via actionable knowledge extraction for embodied tasks, ZSON~\cite{Majumdar2020VLNWebText} using multimodal embeddings for semantic-goal navigation, CLIP-Nav~\cite{Dorbala2022CLIPNav} integrating CLIP for vision-language navigation, and VLFM~\cite{Shah2023VLFM} employing vision-language frontier maps for real-robot deployment. Recent methods incorporate advanced reasoning, such as SG-Nav~\cite{Yin2024SGNav}, which uses scene graph-based reasoning, highlighting trends in generative planning and multi-modal integration. Despite progress, these approaches suffer from modality-specific constraints, simplistic reasoning neglecting spatial-semantic complexities, and inefficient exploration~\cite{Long2023Advances}. For instance, ZSON~\cite{Majumdar2020VLNWebText} and SG-Nav~\cite{Yin2024SGNav} require separate architectures for different goal types, leading to fragmented systems, while unified frameworks like UniGoal~\cite{yin2025unigoal} target universality but underexploit visual properties for instance discrimination, often treating instance images as simple specifications rather than leveraging them for continuous verification and graph-based obstacle avoidance/loop closure. Our framework addresses these gaps by enhancing graph representations with rich visual features and semantic matching, enabling balanced exploration and navigation while reducing redundant paths and hallucinations in a universal zero-shot manner.

%% file: 03_system.tex
\section{System Overview}
We propose $\text{T}^2$-Nav, a zero-shot visual navigation framework that focuses on IIN, integrating two synergistic components: Temporal Graph Memory Networks and Topological Signatures for Loop Closure. Our proposed framework is illustrated in Figure~\ref{fig:pipe}. \del{Our key insight is that navigation patterns can be characterized through both the temporal evolution of scene graphs and topological invariants of trajectories, robust spatial reasoning to avoid inefficient exploration patterns without any learned parameters.}The TeRM maintains a temporal buffer of scene graphs with rich visual feature embeddings. TeRM establishes cross-temporal correspondences that capture instance permanence and visual dynamics. This temporal reasoning capability is particularly crucial for instance-image navigation, where the same object instance may appear dramatically different from various viewpoints. The TSLC computes persistent homology features from trajectory embeddings and detects complex navigation loops through topological signatures that remain invariant to metric distortions and environmental variations. This capability is essential, for instance, for image navigation, where visual similarity between instances can lead to confusion and inefficient exploration patterns.

%% file: 04_methodology.tex
\section{Methodology}
\subsection{Problem Formulation}
Visual navigation in unknown environments requires agents to maintain spatial understanding while exploring efficiently toward goals. Given RGB-D observations $\mathcal{O}_t$ and a goal specification $\mathcal{G}_g$ consisting of a reference image showing the target instance, the agent must generate a sequence of actions $\{a_t\}_{t=1}^T$ to reach the target within $T$ timesteps.

Let $\mathcal{G}_t = (\mathcal{V}_t, \mathcal{E}_t)$ denote the scene graph at time $t$, where $\mathcal{V}_t = \{v_i^t\}_{i=1}^{N_t}$ represents $N_t$ detected objects and $\mathcal{E}_t \subseteq \mathcal{V}_t \times \mathcal{V}_t$ encodes spatial relationships. Each node $v_i^t$ contains attributes $\mathbf{a}_i^t = [\mathbf{p}_i^t, \mathbf{f}_i^t, l_i^t]$ comprising position $\mathbf{p}_i^t \in \mathbb{R}^3$, visual features $\mathbf{f}_i^t \in \mathbb{R}^d$, and $l_i^t$ is semantic label, which is class name of the detected object at node \textit{i} and time \textit{t}. The goal graph $\mathcal{G}_g = (\mathcal{V}_g, \mathcal{E}_g)$ similarly encodes the target configuration.

The agent trajectory up to time $t$ is denoted as $\boldsymbol{\tau}_t = \{(\mathbf{x}_i, \mathbf{y}_i, \theta_i)\}_{i=1}^t$, where $(\mathbf{x}_i, \mathbf{y}_i)$ represents the 2D position and $\theta_i$ the orientation. The exploration policy $\pi: \mathcal{G}_t \times \mathcal{G}_g \times \boldsymbol{\tau}_t \rightarrow \mathcal{A}$ maps the current scene understanding to actions $\mathcal{A} = \{\text{forward}, \text{turn-left}, \text{turn-right}, \text{stop}\}$.

\subsection{Temporal Reasoning Memory Module}
\begin{figure}[ht]
    \centering
    \includegraphics[width=\linewidth]{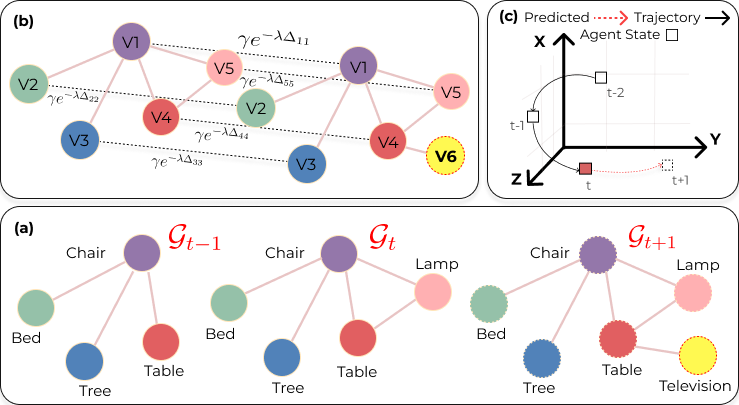}
    \caption{TeRM Framework Overview. (a) Scene graphs track object persistence and new entities across timesteps. (b) Temporal edge construction for dependencies between nodes using exponentially decayed edge weights across $t$. (c) Physics-inspired positions prediction refines object trajectories based on motion dynamics.
    }
    \label{fig:temp}
\end{figure}

To address the challenge of modeling the temporal dynamics in scene understanding while maintaining robust spatial memory, we propose TeRM module, illustrated in Figure~\ref{fig:temp}. This module represent a paradigm shift from static scene representation to a dynamic, temporally-aware graph structures that capture the evolving nature of real-world environments with a sliding window comprising the most recent \( K \) graph snapshots, denoted as \( \{\mathcal{G}_{t-k}\}_{k=0}^{K} \), where each \( \mathcal{G}_{t-k} \) represents the agent's scene graph at timestep \( t-k \). To model the natural temporal decay of information relevance over time, we incorporate a temporal discount factor \( \gamma \in (0,1) \) that progressively reduces the influence of older snapshots while maintaining their contribution to overall scene understanding.

\subsubsection{Cross-Temporal Instance Linking}
The TeRM module link matching semantically and spatially corresponding entities across consecutive temporal snapshots, for every pair of consecutive graph snapshots (\textit{e.g.}, \( \mathcal{G}_{t-1} \) and \( \mathcal{G}_t \)), we establish temporal edges to link semantically and spatially corresponding nodes across time. Specifically, for nodes \( v_i^{t-1} \) in \( \mathcal{G}_{t-1} \) and \( v_j^t \) in \( \mathcal{G}_t \), a temporal edge \( e_{ij}^{\text{temp}} \) is added only if their similarity exceeds $\tau \in [0,1]$ which is the minimum similarity threshold for edge formation for ensuring only reliable correspondences are retained:

\begin{equation}
e_{ij}^{\text{temp}} = \begin{cases}
(\gamma e^{-\lambda \Delta_{ij}}, \Delta t) & \text{if } \text{sim}(v_i^{t-1}, v_j^t) > \tau \\
\varnothing & \text{otherwise},
\end{cases}
\end{equation}
where \( \Delta_{ij} = \|\mathbf{f}_i^{t-1} - \mathbf{f}_j^t\|_2 \) quantifies the Euclidean distance between the visual feature vectors \( \mathbf{f}_i^{t-1} \) and \( \mathbf{f}_j^t \) of the respective nodes, thereby capturing changes in appearance, \( \Delta t \) denotes the discrete timestep difference between the snapshots, and \( \lambda > 0 \) is a tunable hyperparameter governing the exponential decay rate based on appearance variation, ensuring that dramatically different appearances receive appropriately reduced correspondence weights.

\subsubsection{Instance Similarity Computation}
To compute the similarity between a pair of nodes \( v_i \) and \( v_j \), we employ the similarity metric \( \text{sim}(v_i, v_j) \), which fuses semantic label agreement with spatial closeness in a weighted manner:

\begin{equation}
\small
\text{sim}(v_i, v_j) = \alpha \cdot \mathbf{1}[l_i = l_j] + (1-\alpha) \cdot \exp\left(-\frac{\|\mathbf{p}_i - \mathbf{p}_j\|_2}{\sigma_p}\right),
\label{eq1}
\end{equation}
where \( \mathbf{1}[l_i = l_j] \) is an indicator function that equals 1 if the semantic labels \( l_i \) and \( l_j \) of the nodes match (\textit{e.g.}, both represent the same object class) and 0 otherwise, \( \mathbf{p}_i \) and \( \mathbf{p}_j \) denoted of 3D spatial positions of the nodes, \( \alpha \) is an empirically tuned weight that balances semantic matching against spatial alignment, \( \sigma_p \) serves as the standard deviation for the Gaussian spatial decay, controlling tolerance to positional differences. This mechanism enables robust tracking of evolving scene elements while mitigating noise from transient observations.

\subsubsection{Velocity Estimations}
Given the temporal graph structure, TeRM predicts future instance states using velocity extracted from cross-temporal edges. For each tracked instance, we computes velocity as the discrete temporal derivative of position $\mathbf{v}_{l_i} = \mathbf{p}_i^t - \mathbf{p}_i^{t-1} / \Delta t$ for velocity estimates through discrete position extrapolation. For each node in the current scene graph, the system implements a linear extrapolation scheme based on the established velocity patterns, with a node with $l_i$ and current position $\mathbf{P}_i^t$ the predicted position after $k$ discrete time steps is computed by $\hat{\mathbf{P}}_i^{t+k} = \mathbf{P}_i^t + k \cdot \mathbf{v}_{l_i}$. Nodes without recorded velocities maintain their current positions, reflecting the assumption that static objects remain stationary in the absence of observed motion. This predictive capability enables counterfactual reasoning about potential goal locations. Furthermore, the TeRM module also provides temporal context retrieval, allowing the agent to query the historical trajectory of specific objects, for a given semantic label $\ell$, the system can extract a temporal sequence $\{(t_i, \mathbf{p}_i)\}$ of all position observations of that object class within a specified time window, facilitating spatial pattern recognition and behavioral prediction. 

\begin{figure}
    \centering
    \includegraphics[width=\linewidth]{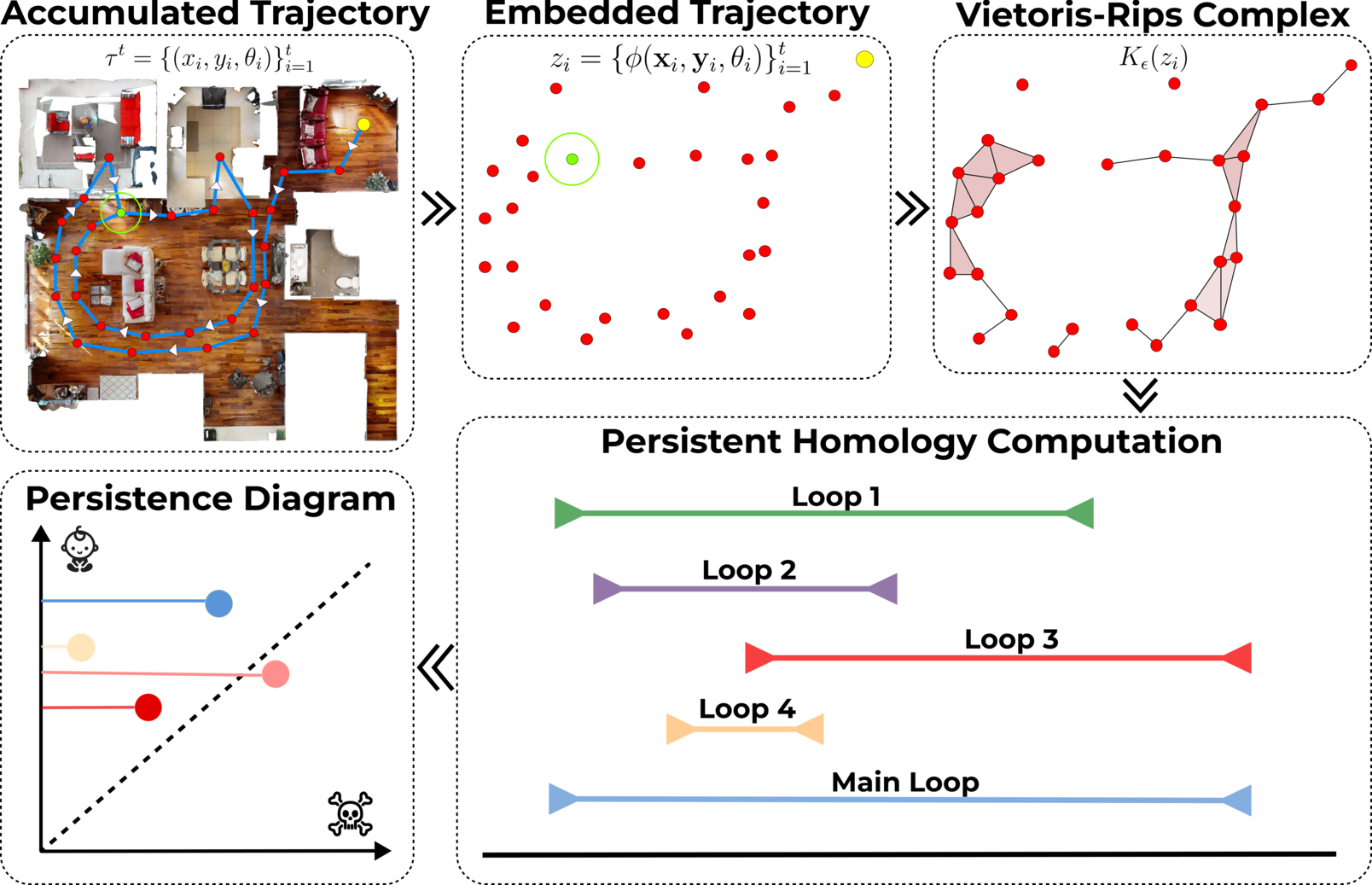}
    \caption{TSLC illustration: Agent trajectories are processed through Vietoris-Rips filtration to derive persistent homological features, which are subsequently mapped to a persistence diagram in a birth-death coordinate system. Loop closure is identified by assessing whether the $W_2$ distance between the persistence diagram of the current trajectory segment and those of historical segments falls below a predefined threshold.
    }
    \label{fig:topo}
\end{figure}

\subsection{Topological Signatures for Loop Closure}
This module aims to address fundamental limitations in current training-free navigation systems by using persistent homology to detect complex loop patterns that go beyond simple geometric proximity. While baseline approaches \cite{yin2025unigoal} rely on pairwise visual feature matching or basic geometric constraints, our method captures intrinsic topological invariants of robot trajectories that remain stable under metric distortions, orientation changes, and environmental noise. The core insight driving our design is that navigation loops manifest as persistent homological features in the trajectory's underlying simplicial structure. For clarity, Figure~\ref{fig:topo} provides an overview of the TSLC components.


\subsubsection{Trajectory Embedding and Complex Construction}
Given an accumulated robot trajectory $\tau^t$ comprising pose sequences with positional coordinates $(x_i, y_i)$ and orientation $\theta_i$, we project each waypoint into an augmented three-dimensional feature space that explicitly integrates both spatial and directional information. The embedding function $\phi: \mathbb{R}^2 \times S^1 \rightarrow \mathbb{R}^3$ for the $i$-th trajectory point is defined as:

\begin{equation}
z_i = \phi((x_i, y_i, \theta_i)) = [x_i, y_i, r \sin(\theta_i)] \in \mathbb{R}^3
\end{equation}
where the scalar parameter $r = 0.5$ empirically modulates the relative influence of the orientation component. The sine projection $\sin(\theta_i)$ encodes heading periodicity while avoiding discontinuities inherent in direct angular representation, ensuring balanced contribution of directional information without overwhelming the spatial dimensions. This embedding preserves the essential geometric structure of the trajectory while enabling subsequent topological analysis through simplicial complex construction. Subsequently, we construct a Vietoris-Rips simplicial complex $K_\epsilon(\tau^t)$ over the embedded point cloud $\{z_i\}_{i=1}^t$, the complex at scale parameter $\epsilon \geq 0$ is defined by:

\begin{equation}
K_\epsilon(\tau^t) = \{\sigma \subseteq \{z_i\}_{i=1}^t : \text{diam}(\sigma) \leq \epsilon\}
\end{equation}
where $\sigma$ denotes arbitrary simplices (subsets of points) and $\text{diam}(\sigma) = \max_{u,v \in \sigma} \|u - v\|_2$ represents the diameter under Euclidean distance. We set the maximum edge length $\epsilon_{\max} = 5.0$ meters and construct the complex up to dimension 2. The construction proceeds by first computing pairwise distances $d_{ij} = \|z_i - z_j\|_2$ for all point pairs, then including edges $(i,j)$ where $d_{ij} \leq \epsilon$, and finally adding triangles $(i,j,k)$ where all three pairwise distances satisfy the constraint. By varying $\epsilon$ across the range $[0, \epsilon_{\max}]$, we obtain a filtration $\{K_\epsilon\}_{\epsilon \geq 0}$ that systematically reveals the evolving topological connectivity of the trajectory at multiple scales.

\subsubsection{Persistent Homology Computation}
Building upon the multi-scale filtration $\{K_\epsilon\}_{\epsilon \geq 0}$, we compute persistent homology groups $H_k(K_\epsilon)$ for each filtration level and homological dimension $k$. The homology groups are defined as the quotient of kernel and image spaces of boundary operators as in follow:

\begin{equation}
H_k(K_\epsilon) = \frac{\text{ker}(\partial_k: C_k(K_\epsilon) \rightarrow C_{k-1}(K_\epsilon))}{\text{im}(\partial_{k+1}: C_{k+1}(K_\epsilon) \rightarrow C_k(K_\epsilon))}
\end{equation}
where $C_k(K_\epsilon)$ denotes the $k$-th chain group over the field $\mathbb{F}_2$, and $\partial_k$ represents the $k$-th boundary operator. The boundary operator for a $k$-simplex $\sigma = [v_0, v_1, \ldots, v_k]$:

\begin{equation}
\partial_k(\sigma) = \sum_{i=0}^k (-1)^i [v_0, \ldots, \hat{v_i}, \ldots, v_k],
\label{eqpar}
\end{equation}
where $\hat{v_i}$ indicates the omission of vertex $v_i$. The persistence computation employs matrix reduction techniques on the boundary matrix $\partial = [\partial_1 | \partial_2 | \cdots | \partial_{\text{max}}]$, where each column represents a simplex ordered by filtration value. The reduction process applies elementary column operations to reach reduced echelon form, in which each column contains at most one pivot (the lowest non-zero entry).

For loop-closure detection, we emphasize 1-dimensional homological features corresponding to cycles in the trajectory topology. These are encapsulated in the persistence diagram for dimension 1, denoted $\text{PD}_1(\tau^t)$, comprising birth-death pairs $(b_i, d_i)$ where a loop emerges at scale $b_i$ and disappears at scale $d_i$:

\begin{equation}
\text{PD}_1(\tau^t) = \{(b_i, d_i) : H_1(K_{b_i}) \neq H_1(K_{d_i})\}
\end{equation}

To distinguish salient topological signals from ephemeral artifacts arising from sampling irregularities, we apply a persistence threshold $\tau_p > 0$, retaining only features where persistence $p_i = d_i - b_i > \tau_p$. This filtering mechanism ensures extracted loops reflect meaningful revisits or cyclic patterns rather than noise-induced fluctuations.

\subsubsection{Topological Signature Matching}
Leveraging persistence diagrams as compact topological descriptors, we enable loop closure detection by quantifying similarity between current trajectory segments $\tau^t$ and historical segments $\tau^s$ stored from previous exploration phases. We employ the 2-Wasserstein distance to measure optimal transport cost between Persistence Diagrams (PD) by $W_p(\text{PD}_1(\tau^t), \text{PD}_1(\tau^s)) = \left(\inf_{\gamma \in \Gamma} \int_{\del{\mathbb{R}^2 \times \mathbb{R}^2} } \|x - y\|_p^p \, d\gamma(x,y)\right)^{\frac{1}{p}}$, then the 2-Wasserstein distance is then computed as:

\begin{equation}
W_2(\text{PD}_1(\tau^t), \text{PD}_1(\tau^s)) = \left(\sum_{(i,j) \in \Pi^*} C_{ij}\right)^{1/2} \end{equation}
where $\Pi^*$ represents the optimal matching obtained via the Hungarian algorithm and $C_{ij}$ denotes the cost matrix entries. This metric is stable under small perturbations of the underlying point cloud and accounts for both the positions and multiplicities of persistent features.

A potential loop closure is flagged when the computed distance falls below an empirically determined threshold $\theta_W$, triggering geometric verification and integration into the agent's global map. The Wasserstein distance computation utilizes the Hungarian algorithm for optimal matching, ensuring both computational efficiency and theoretical optimality guarantees.

\subsubsection{Persistence Landscape Representation}
While persistence diagrams provide a robust topological summary, their multiset nature lacks a direct vector-space structure, which complicates their integration with statistical tools or machine-learning pipelines for tasks such as averaging multiple diagrams or kernel-based classification. To address this, we transform the raw persistence diagrams into persistence landscapes \cite{Bubenik2012StatisticalTD}, a functional representation that embeds topological information into a stable vector space, enabling operations such as computing sample means, distances via $L^p$ norms, or classifications while maintaining Lipschitz continuity for robustness against noise-properties that also contribute to computational efficiency in large-scale or real-time scenarios compared to direct Wasserstein computations. 

For the 1-dimensional persistence diagram $\text{PD}_1(\boldsymbol{\tau}^t)$, the landscape function $\Lambda: \mathbb{R} \to \mathbb{R}$ is defined pointwise as:

\begin{equation}
\Lambda(t) = \max_{(b_i, d_i) \in \text{PD}_1(\boldsymbol{\tau}^t)} \min(t - b_i, d_i - t)^+
\end{equation}
Where $(\cdot)^+$ denotes the positive part (\textit{i.e.}, $\max(\cdot, 0)$), effectively stacking ``tent'' functions centered at the midpoints of each birth-death interval and scaled by their persistence. This layered summary of prominent features (with higher layers for longer-persisting loops) can be discretized over a grid or converted into fixed-length vectors, augmenting or replacing Wasserstein-based matching in resource-constrained applications while preserving the topological essence derived from the trajectory embedding, filtration, and homology computation.

\subsubsection{Multi-Modal Feature Integration}
To enhance topological discrimination, our method supports integration of visual features extracted from RGB observations. Given pre-trained visual encoders producing feature vectors $f_i \in \mathbb{R}^d$ at each trajectory point, we construct an enriched embedding:

\begin{equation}
z_i^{\text{enhanced}} = [x_i, y_i, r \sin(\theta_i)] \oplus \alpha \cdot \text{SVD}_3(f_i)
\end{equation}
Where $\text{SVD}_3(\cdot)$ denotes projection to the first three singular vectors via SVD decomposition, $\oplus$ represents concatenation, and $\alpha > 0$ weights the visual contribution. This multi-modal embedding preserves spatial-temporal structure while incorporating visual distinctiveness, yielding more discriminative topological signatures without compromising the approach's training-free nature. Algorithm~\ref{alg:tslc} provides the procedure for the TSLC module, incorporating these multi-modalities features.

\begin{algorithm}[!t]
\footnotesize
\caption{TSLC Pseudo-Algorithms}
\label{alg:tslc}
\begin{algorithmic}[1]
\Require $\tau^t = \{(x_i, y_i, \theta_i)\}_{i=1}^t$, $\mathcal{S} = \{(\mathcal{T}_j, \mathbf{p}_j)\}_{j=1}^{|\mathcal{S}|}$, $\theta_W$
\Ensure Loop Detected, Matched Index, Confidence
\If{$|\tau^t| < 10$} \Return $(0, \varnothing, 0.0)$ \EndIf
\State $\mathcal{Z} \gets \{[x_i, y_i, r\sin(\theta_i)]\}_{i=1}^t$
\State Construct Vietoris-Rips complex with $\epsilon_{\max} = 5.0$
\State Compute $\text{PD}_1$ via Union-Find persistent homology
\State $\text{PD}_1^f \gets \{(b,d) \in \text{PD}_1 : d-b > 0.1\}$
\State $\Lambda(t) \gets \max_{(b,d) \in \text{PD}_1^f} \min(t-b, d-t)^+$
\State $\mathcal{T} \gets \{\text{PD}_1^f, \Lambda\}$
\For{$j \in \{1, \ldots, |\mathcal{S}|\}$}
    \If{$\|\mathbf{p}_{curr} - \mathbf{p}_j\|_2 > R_{search}$} \textbf{continue} \EndIf
    \State $d \gets 0.7 \cdot W_2(\text{PD}_1^f, \text{PD}_1^f(\tau^j)) + 0.3 \cdot \|\Lambda - \Lambda_j\|_2$
    \State Update best match if $d < d_{min}$
\EndFor
\State $\text{detected} \gets \mathbb{I}[d_{min} < \theta_W]$
\State $\mathcal{S} \gets \mathcal{S} \cup \{(\mathcal{T}, \mathbf{p}_{curr})\}$
\State \Return $(\text{detected}, j^*, \exp(-d_{min}/\theta_W))$
\end{algorithmic}
\end{algorithm}


%% file: 05_evaluation.tex
\section{Evaluations}
\subsection{Datasets and Experiments Setup}

\textbf{Datasets:} We use the HM3D dataset \cite{ramakrishnan2021hm3d} in the Habitat 2.0 simulator, which contains 1,000 high-resolution indoor reconstructions with object-level annotations. For the IIN task, the agent must reach a target object instance given a goal image, requiring visual matching and efficient path planning, following the setup of Krantz \textit{et al.} \cite{krantz2022instance}.

\textbf{Evaluation Metrics}: We evaluate our method using two standard metrics for embodied navigation tasks, as in UniGoal \cite{yin2025unigoal}, with Success Rate (SR) to measure the percentage of episodes where the agent successfully reaches the target within the distance threshold. An episode is successful if the agent terminates within 1.0 meters of the target object instance, and Success is weighted by Path Length (SPL)  to evaluate both task completion and navigation efficiency. This metric can be expressed by Equation~\eqref{eq3}.
\begin{equation}
    \text{SPL} = \frac{1}{N} \sum_{i=1}^{N} S_i \cdot \frac{\max(p_i, l_i)}{l_i}
\label{eq3}
\end{equation}
where $N$ and $S_i$ represent the total number of episodes and a binary success indicator for each episode, respectively. $i$, $l_i$ is the shortest path distance from the start to the goal position, and $p_i$ is the path length traversed by the agent. Higher SPL value indicates more efficient navigation, as it penalizes agents that take unnecessarily long paths to reach the target.

\textbf{Environment}: We use NetworkX for graph operations and GUDHI for persistent homology computation. The temporal memory maintains $K = 100$ snapshots with decay factor $\gamma = 0.95$. For TSLC, we use a maximum edge length $\epsilon_{\max} = 5.0$ meters and compute persistence up to dimension 2, though only 1-dimensional features are used for loop detection. The persistence threshold $\tau_p = 0.1$ filters noise-induced features, while the Wasserstein threshold $\theta_W = 2.0$ determines loop closure sensitivity. Topological signatures are computed every 10 time steps to balance computational efficiency with detection sensitivity. The orientation weight parameter $r = 0.5$ ensures balanced contribution of directional information without overwhelming spatial coordinates. For the object detection and scene graph construction pipeline, we follow UniGoal \cite{yin2025unigoal}, using GroundingDINO \cite{gdinob} for open-vocabulary instance segmentation and LLaMA as the LLM, LLaVA as the VLM, and CLIP text encoder for node and edge embeddings within the graph matching module. Spatial relationships between detected objects are proposed via an LLM query. All experiments were conducted on a single NVIDIA A6000 GPU.

\subsection{Quantitative Results}

\begin{table}[ht]
\centering
\begin{tabular}{l c c c c}
\toprule
\textbf{Method} & \textbf{Training-Free} & \textbf{Input} & \textbf{SR $\uparrow$} & \textbf{SPL $\uparrow$} \\
\midrule
Krantz \textit{et al.} \cite{krantz2022instance} & \textcolor{red}{$\times$} & RGBD$^{++}$ & 8.3 & 3.5 \\
ZSON \cite{majumdar2022zson} & \textcolor{red}{$\times$} & RGB$^{+}$ & 14.6 & 7.3 \\
OVRL-v2 \cite{yadav2023ovrlv2} & \textcolor{red}{$\times$} & RGB$^{++}$ & 24.8 & 11.8 \\
IEVE \cite{lei2024instance} & \textcolor{red}{$\times$} & RGBD$^{++}$ & 70.2 & 25.2 \\
PSL \cite{sun2024prioritized} & \textcolor{red}{$\times$} & RGB$^{-}$ & 23.0 & 11.4 \\
GOAT \cite{chang2023goat} & \textcolor{red}{$\times$} & RGBD & 37.4 & 16.1 \\
\midrule
Mod-IIN \cite{krantz2023navigating} & \textcolor{green}{\checkmark} & RGBD$^{++}$ & 56.1 & 23.3 \\
UniGoal \cite{yin2025unigoal} & \textcolor{green}{\checkmark} & RGBD$^+$ & 60.2 & 23.7 \\
\textbf{$\text{T}^2$-Nav} & \textcolor{green}{\checkmark} & RGBD$^{++}$ & \textbf{72.6} & \textbf{27.8} \\
\bottomrule
\end{tabular}
\caption{Instance-image-goal navigation results on HM3D, methods with training-free capability are marked with \textcolor{green}{\checkmark}, where $^{-}$, $^{+}$, and $^{++}$ denote only orientation input, only coordinate input, and for both of them, respectively.}
\label{tab:navigation_results}
\end{table}

The performance of T$^{2}$-Nav is outlined in Table~\ref{tab:navigation_results}, compared with recent methods on the HM3D IIN task. On this benchmark, T$^{2}$-Nav surpasses the baseline results of UniGoal with a $+12.4/4.1$ lead on SR/SPL. It also outperforms the best supervised method, IEVE, with a $+2.4/2.6$ lead on SR/SPL, even without any task-specific training. These results show that the combination of TeRM and TSLC brings clear gains, the method performs well without learning parameters, and still provides both higher success and more efficient paths than supervised or training-free baselines.

\subsection{Qualitative Results}

\begin{figure}[!t]
    \centering
    \includegraphics[width=\linewidth]{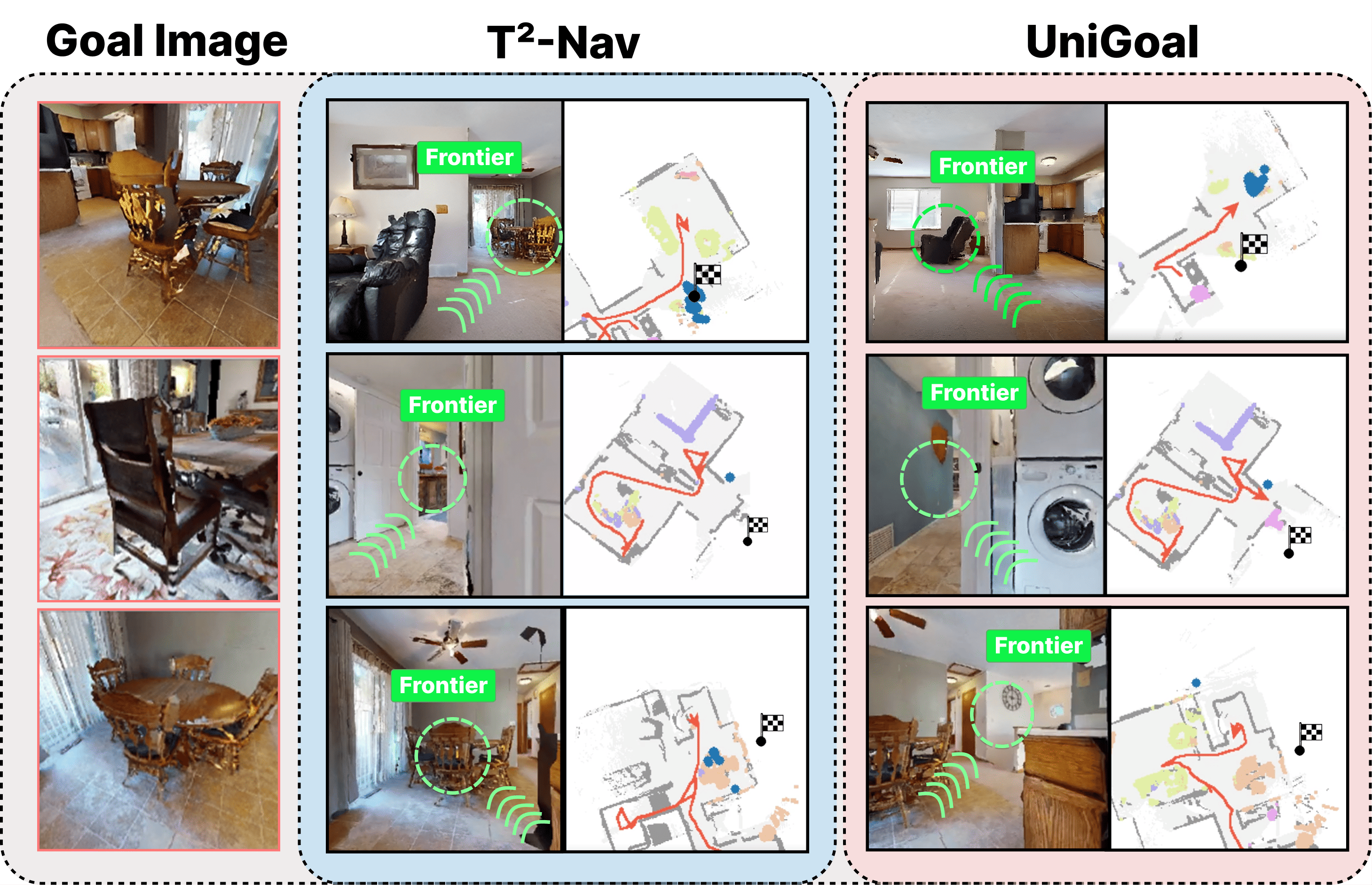}
    \caption{Qualitative comparison of frontier selection strategies between UniGoal and the proposed method. Compared with the baseline UniGoal, $\text{T}^2$-Nav selects frontiers more strategically, for more efficient exploration and reducing redundant path traversals.}
    \label{fig:frontier}
\end{figure}

Figure~\ref{fig:frontier} presents representative navigation episodes comparing T$^{2}$-Nav with the UniGoal baseline; the proposed T$^{2}$-Nav consistently selects frontiers that guide exploration toward unexplored regions likely to contain the goal, thereby avoiding unnecessary detours. In contrast, UniGoal often expands frontiers in a less structured manner, leading to longer or redundant paths before reaching the target, and sometimes even passing the target without recognizing it. These observations showed that the temporal reasoning and topological loop-detection components not only enhance the quantitative metrics but also yield more reliable, goal-directed behavior in complex and cluttered environments.

\begin{figure}[!t]
    \centering
    \includegraphics[width=1\linewidth]{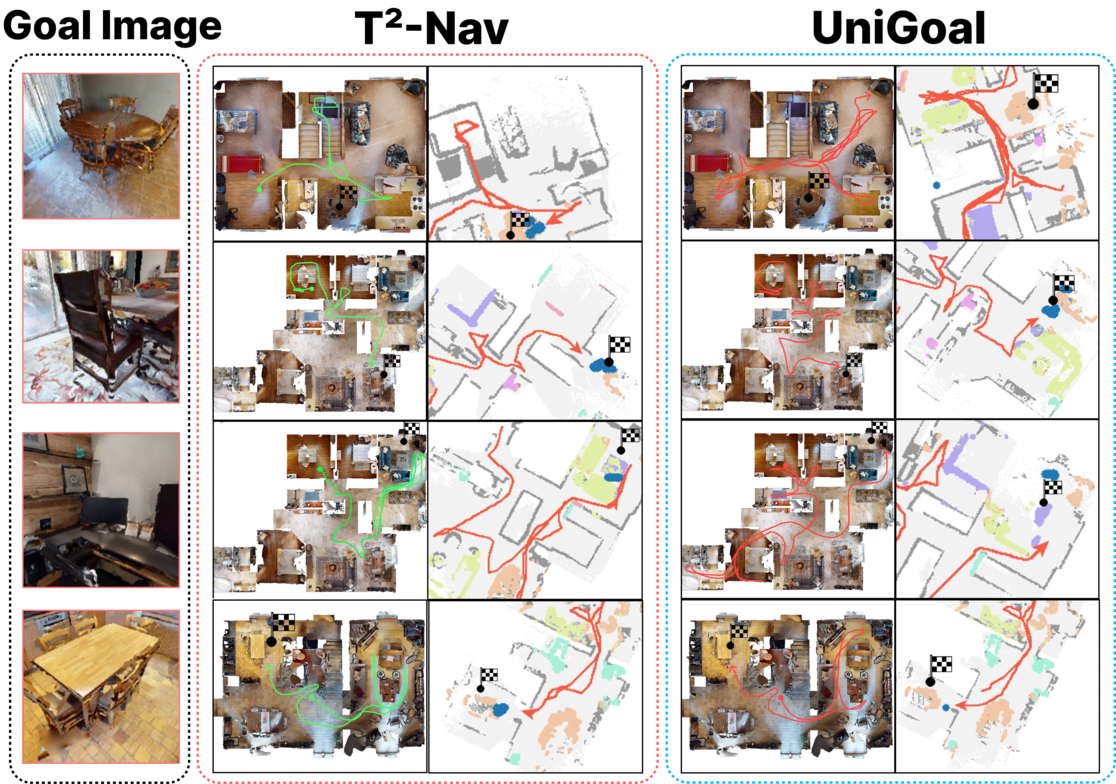}
    \caption{Qualitative comparison of navigation trajectories between UniGoal and the proposed method. Compared with the baseline UniGoal, T$^{2}$-Nav produces shorter, more efficient trajectories with fewer detours, thereby demonstrating improved goal-directed navigation.}
    \label{fig:trajectory}
\end{figure}


Similarly, Figure~\ref{fig:trajectory} showcases the navigation paths by T$^{2}$-Nav compared to UniGoal. The proposed approach consistently produces more concise routes, leading the agent through goal-relevant regions with minimal backtracking. By incorporating temporal reasoning and early loop detection, T$^{2}$-Nav avoids unnecessary wandering and efficiently converges on the goal. Meanwhile, UniGoal tends to take longer, more circuitous routes, often circling through large areas of the environment before reaching the target. In several cases, it overshoots the goal or passes near it without recognizing it, leading to redundant exploration. These trajectory-level comparisons emphasize that T$^{2}$-Nav provides not only quantitative gains but also qualitatively smoother and more goal-directed navigation behavior. Additional experiments are available in the video \href{https://youtu.be/joRCrhP5KxA}{https://youtu.be/joRCrhP5KxA}.

\subsection{Ablation Study}
We present ablation studies of T$^{2}$-Nav to validate the robustness of each component on 200 episodes from the HM3D dataset, as shown in Table~\ref{tab:ab}. Starting from the UniGoal baseline with SR results of 66.16/24.83, adding the TeRM module improves both metrics, yielding gains of $+8.8$ in SR and $+3.5$ in SPL. This highlights the importance of temporal consistency in maintaining scene understanding. Similarly, incorporating the TSLC module boosts performance over the baseline by $+6.1$ SR / $+0.7$ SPL, demonstrating its role in avoiding repeated exploration of previously visited regions. When both components are combined, T$^{2}$-Nav achieves the best performance, with SR and SPL of 75.62/28.38, confirming that TeRM and TSLC are complementary. TeRM enhances goal-directed reasoning over time, while TSLC ensures efficient path selection, together producing the most reliable and efficient navigation.

\begin{table}[h]
\centering
\setlength{\tabcolsep}{10pt} 
\renewcommand{\arraystretch}{1.2} 
\begin{tabular}{l r r}
\toprule
\textbf{Method} & \textbf{SR $\uparrow$} & \textbf{SPL $\uparrow$} \\
\midrule
Baseline \cite{yin2025unigoal} & 66.16 & 24.83 \\
\textit{w/o} Temporal Reasoning Memory Module & 74.99 & 28.28 \\
\textit{w/o} Topological Signatures for Loop Closure & 72.22 & 25.48 \\
\midrule
\textbf{Proposed $\text{T}^2$-Nav} & \textbf{75.62} & \textbf{28.38} \\
\bottomrule
\end{tabular}
\caption{Ablation results on IIN over 200 episodes on HM3D.}
\label{tab:ab}
\end{table}

%% file: 06_conclusions.tex


\section{Conclusions}
In this research, we introduced T$^2$-Nav, a training-free visual navigation framework that utilized two proposed modules, TeRM and TSLC, to achieve robust IIN. Experiments on HM3D show that T$^2$-Nav outperforms both training-free and supervised baselines, highlighting the effectiveness of combining temporal memory and topological reasoning for scalable embodied navigation. While our method excels in indoor environments, extending it to outdoor terrains and multi-floor buildings presents opportunities for using semantic hierarchies to enhance goal understanding. A key limitation remains real-time deployment: the computational overhead of VLM and LLM inference, shared across all foundation-model-based navigation approaches, including our baseline, renders the full pipeline ill-suited for on-robot real-time operation. Future work will explore advanced topological methods beyond persistent homology, lightweight approximations to reduce inference cost, and real-robot deployment to validate generalizability.

%% file: 00_main.bbl
\begin{thebibliography}{10}
\providecommand{\url}[1]{#1}
\csname url@rmstyle\endcsname
\providecommand{\newblock}{\relax}
\providecommand{\bibinfo}[2]{#2}
\providecommand\BIBentrySTDinterwordspacing{\spaceskip=0pt\relax}
\providecommand\BIBentryALTinterwordstretchfactor{4}
\providecommand\BIBentryALTinterwordspacing{\spaceskip=\fontdimen2\font plus
\BIBentryALTinterwordstretchfactor\fontdimen3\font minus \fontdimen4\font\relax}
\providecommand\BIBforeignlanguage[2]{{%
\expandafter\ifx\csname l@#1\endcsname\relax
\typeout{** WARNING: IEEEtran.bst: No hyphenation pattern has been}%
\typeout{** loaded for the language `#1'. Using the pattern for}%
\typeout{** the default language instead.}%
\else
\language=\csname l@#1\endcsname
\fi
#2}}

\bibitem{ZhengYSZWZZC25}
Y.~Zheng, L.~Yao, Y.~Su, Y.~Zhang, Y.~Wang, S.~Zhao, Y.~Zhang, and L.~Chau, ``A survey of embodied learning for object-centric robotic manipulation,'' \emph{Mach. Intell. Res.}, vol.~22, no.~4, pp. 588--626, 2025.

\bibitem{krantz2022instance}
J.~Krantz, S.~Lee, J.~Malik, D.~Batra, and D.~S. Chaplot, ``Instance-specific image goal navigation: Training embodied agents to find object instances,'' \emph{arXiv preprint arXiv:2211.15876}, 2022.

\bibitem{bhattacharya2015persistent}
S.~Bhattacharya, R.~Ghrist, and V.~Kumar, ``Persistent homology for path planning in uncertain environments,'' \emph{IEEE Transactions on Robotics}, vol.~31, no.~3, pp. 578--590, 2015.

\bibitem{pokorny2016high}
F.~T. Pokorny, D.~Kragic, L.~E. Kavraki, and K.~Goldberg, ``High-dimensional winding-augmented motion planning with 2d topological task projections and persistent homology,'' in \emph{2016 IEEE International Conference on Robotics and Automation (ICRA)}, 2016, pp. 24--31.

\bibitem{kwon2021visual}
O.~Kwon, N.~Kim, Y.~Choi, H.~Yoo, J.~Park, and S.~Oh, ``Visual graph memory with unsupervised representation for visual navigation,'' in \emph{Proceedings of the IEEE/CVF international conference on computer vision}, 2021, pp. 15\,890--15\,899.

\bibitem{cai2024dgmem}
W.~Cai, T.~Wang, G.~Cheng, L.~Xu, and C.~Sun, ``Dgmem: learning visual navigation policy without any labels by dynamic graph memory,'' \emph{Applied Intelligence}, vol.~54, no.~17, pp. 8442--8453, 2024.

\bibitem{lei2024gaussnav}
X.~Lei, M.~Wang, W.~Zhou, and H.~Li, ``Gaussnav: Gaussian splatting for visual navigation,'' \emph{{IEEE} Trans. Pattern Anal. Mach. Intell.}, vol.~47, no.~5, pp. 4108--4121, 2025.

\bibitem{Xiao2025}
F.~Xiao, Y.~Zhang, J.~Fang, X.~Guo, and R.~Huang, ``Improving the navigation optimization of hospital logistics robots under complex lighting changes by using improved orb-slam3 and deep learning visual slam algorithm,'' \emph{Discover Applied Sciences}, vol.~7, no.~4, p. 291, Apr 2025.

\bibitem{SUN2025103135}
L.~Sun, A.~Kanezaki, G.~Caron, and Y.~Yoshiyasu, ``Enhancing multimodal-input object goal navigation by leveraging large language models for inferring room–object relationship knowledge,'' \emph{Advanced Engineering Informatics}, vol.~65, p. 103135, 2025.

\bibitem{Wijmans2020DDPPO}
E.~Wijmans, A.~Kadian, A.~S. Morcos, S.~Lee, I.~Essa, D.~Parikh, M.~Savva, and D.~Batra, ``Dd-ppo: Learning near-perfect pointgoal navigators from 2.5 billion frames,'' in \emph{International Conference on Learning Representations (ICLR)}, 2020.

\bibitem{Anderson2018VLN}
P.~Anderson, Q.~Wu, D.~Teney, J.~Bruce, M.~Johnson, N.~S{\"u}nderhauf, I.~Reid, S.~Gould, and A.~van~den Hengel, ``Vision-and-language navigation: Interpreting visually-grounded navigation instructions in real environments,'' in \emph{Proceedings of the IEEE/CVF Conference on Computer Vision and Pattern Recognition (CVPR)}, 2018, pp. 3674--3683.

\bibitem{Ramakrishnan_2022_CVPR}
S.~K. Ramakrishnan, D.~S. Chaplot, Z.~Al-Halah, J.~Malik, and K.~Grauman, ``Poni: Potential functions for objectgoal navigation with interaction-free learning,'' in \emph{Proceedings of the IEEE/CVF Conference on Computer Vision and Pattern Recognition (CVPR)}, June 2022, pp. 18\,890--18\,900.

\bibitem{Dorbala2022CLIPNav}
V.~S. Dorbala, G.~Sigurdsson, R.~Piramuthu, J.~Thomason, and G.~S. Sukhatme, ``Clip-nav: Using clip for zero-shot vision-and-language navigation,'' \emph{arXiv preprint arXiv:2211.16649}, 2022.

\bibitem{pmlr-v205-shah23b}
D.~Shah, B.~Osiński, B.~Ichter, and S.~Levine, ``Lm-nav: Robotic navigation with large pre-trained models of language, vision, and action,'' in \emph{Proceedings of The 6th Conference on Robot Learning}, ser. Proceedings of Machine Learning Research, K.~Liu, D.~Kulic, and J.~Ichnowski, Eds., vol. 205.\hskip 1em plus 0.5em minus 0.4em\relax PMLR, 14--18 Dec 2023, pp. 492--504.

\bibitem{Majumdar2020VLNWebText}
A.~Majumdar, A.~Shrivastava, S.~Lee, P.~Anderson, D.~Parikh, and D.~Batra, ``Improving vision-and-language navigation with image-text pairs from the web,'' in \emph{Computer Vision -- ECCV 2020, Lecture Notes in Computer Science, Vol. 12351}, 2020, pp. 259--274.

\bibitem{Yin2024SGNav}
H.~Yin, X.~Xu, Z.~Wu, J.~Zhou, and J.~Lu, ``Sg-nav: online 3d scene graph prompting for llm-based zero-shot object navigation,'' in \emph{Proceedings of the 38th International Conference on Neural Information Processing Systems}, ser. NIPS '24.\hskip 1em plus 0.5em minus 0.4em\relax Curran Associates Inc., 2025.

\bibitem{yin2025unigoal}
H.~Yin, X.~Xu, L.~Zhao, Z.~Wang, J.~Zhou, and J.~Lu, ``Unigoal: Towards universal zero-shot goal-oriented navigation,'' in \emph{{IEEE/CVF} Conference on Computer Vision and Pattern Recognition, {CVPR} 2025, Nashville, TN, USA, June 11-15, 2025}.\hskip 1em plus 0.5em minus 0.4em\relax Computer Vision Foundation / {IEEE}, 2025, pp. 19\,057--19\,066.

\bibitem{pmlr-v162-li22n}
J.~Li, D.~Li, C.~Xiong, and S.~Hoi, ``{BLIP}: Bootstrapping language-image pre-training for unified vision-language understanding and generation,'' in \emph{Proceedings of the 39th International Conference on Machine Learning}, ser. Proceedings of Machine Learning Research, K.~Chaudhuri, S.~Jegelka, L.~Song, C.~Szepesvári, G.~Niu, and S.~Sabato, Eds., vol. 162.\hskip 1em plus 0.5em minus 0.4em\relax PMLR, 17--23 Jul 2022, pp. 12\,888--12\,900.

\bibitem{Zhu2017TargetDriven}
Y.~Zhu, R.~Mottaghi, E.~Kolve, J.~L. Joseph\, A.~Gupta, L.~Fei-Fei, and A.~Farhadi, ``Target-driven visual navigation in indoor scenes using deep reinforcement learning,'' in \emph{Proceedings of the IEEE International Conference on Robotics and Automation (ICRA)}, July 2017, pp. 3357--3364.

\bibitem{Feng2025SurveyLLM}
J.~Feng, J.~Zeng, Q.~Long, H.~Chen, J.~Zhao, Y.~Xi, Z.~Zhou, Y.~Yuan, S.~Wang, Q.~Zeng, S.~Li, Y.~Zhang, Y.~Lin, T.~Li, J.~Ding, C.~Gao, F.~Xu, and Y.~Li, ``A survey of large language model-powered spatial intelligence across scales: Advances in embodied agents, smart cities, and earth science,'' \emph{arXiv preprint arXiv:2504.09848}, 2025.

\bibitem{Shah2023VLFM}
N.~Yokoyama, S.~Ha, D.~Batra, J.~Wang, and B.~Bucher, ``{VLFM:} vision-language frontier maps for zero-shot semantic navigation,'' in \emph{{IEEE} International Conference on Robotics and Automation, {ICRA} 2024, Yokohama, Japan, May 13-17, 2024}, 2024, pp. 42--48.

\bibitem{Long2023Advances}
J.~Lin, H.~Gao, X.~Feng, R.~Xu, C.~Wang, M.~Zhang, L.~Guo, and S.~Xu, ``The development of llms for embodied navigation,'' \emph{CoRR}, vol. abs/2311.00530, 2023.

\bibitem{Bubenik2012StatisticalTD}
P.~Bubenik, ``Statistical topological data analysis using persistence landscapes,'' \emph{J. Mach. Learn. Res.}, vol.~16, pp. 77--102, 2012.

\bibitem{ramakrishnan2021hm3d}
S.~K. Ramakrishnan, A.~Gokaslan, E.~Wijmans, O.~Maksymets, A.~Clegg, J.~M. Turner, E.~Undersander, W.~Galuba, A.~Westbury, A.~X. Chang, M.~Savva, Y.~Zhao, and D.~Batra, ``Habitat-matterport 3d dataset ({HM}3d): 1000 large-scale 3d environments for embodied {AI},'' in \emph{Thirty-fifth Conference on Neural Information Processing Systems Datasets and Benchmarks Track (Round 2)}, 2021.

\bibitem{gdinob}
S.~Liu, Z.~Zeng, T.~Ren, F.~Li, H.~Zhang, J.~Yang, Q.~Jiang, C.~Li, J.~Yang, H.~Su, J.~Zhu, and L.~Zhang, ``Grounding dino: Marrying dino with grounded pre-training for open-set object detection,'' in \emph{Computer Vision -- ECCV 2024}, A.~Leonardis, E.~Ricci, S.~Roth, O.~Russakovsky, T.~Sattler, and G.~Varol, Eds.\hskip 1em plus 0.5em minus 0.4em\relax Cham: Springer Nature Switzerland, 2025, pp. 38--55.

\bibitem{majumdar2022zson}
A.~Majumdar, G.~Aggarwal, B.~Devnani, J.~Hoffman, and D.~Batra, ``Zson: Zero-shot object-goal navigation using multimodal goal embeddings,'' in \emph{Advances in Neural Information Processing Systems (NeurIPS)}, vol.~35, 2022, pp. 32\,340--32\,352.

\bibitem{yadav2023ovrlv2}
K.~Yadav, A.~Majumdar, R.~Ramrakhya, N.~Yokoyama, A.~Baevski, Z.~Kira, O.~Maksymets, and D.~Batra, ``Ovrl-v2: A simple state-of-art baseline for imagenav and objectnav,'' \emph{arXiv preprint arXiv:2303.07798}, 2023.

\bibitem{lei2024instance}
X.~Lei, M.~Wang, W.~Zhou, L.~Li, and H.~Li, ``Instance-aware exploration-verification-exploitation for instance imagegoal navigation,'' in \emph{Proceedings of the IEEE/CVF Conference on Computer Vision and Pattern Recognition (CVPR)}, 2024, pp. 16\,329--16\,339.

\bibitem{sun2024prioritized}
X.~Sun, L.~Liu, H.~Zhi, R.~Qiu, and J.~Liang, ``Prioritized semantic learning for zero-shot instance navigation,'' in \emph{Computer Vision -- ECCV 2024}, A.~Leonardis, E.~Ricci, S.~Roth, O.~Russakovsky, T.~Sattler, and G.~Varol, Eds.\hskip 1em plus 0.5em minus 0.4em\relax Cham: Springer Nature Switzerland, 2025, pp. 161--178.

\bibitem{chang2023goat}
M.~Chang, T.~Gervet, M.~Khanna, S.~Yenamandra, D.~Shah, S.~Y. Min, K.~Shah, C.~Paxton, S.~Gupta, D.~Batra, \emph{et~al.}, ``Goat: Go to any thing,'' \emph{arXiv preprint arXiv:2311.06430}, 2023.

\bibitem{krantz2023navigating}
J.~Krantz, T.~Gervet, K.~Yadav, A.~Wang, C.~Paxton, R.~Mottaghi, D.~Batra, J.~Malik, S.~Lee, and D.~S. Chaplot, ``Navigating to objects specified by images,'' in \emph{Proceedings of the IEEE/CVF International Conference on Computer Vision (ICCV)}.\hskip 1em plus 0.5em minus 0.4em\relax IEEE, 2023, pp. 10\,916--10\,925.

\end{thebibliography}
